\documentclass[11pt,twoside,a4paper]{article}
\usepackage{pgm}
\usepackage{graphics}
\usepackage{amsthm}

\theoremstyle{definition}
\theoremstyle{remark}
\newcommand{\doo}{\mbox{do}}
\newcommand{\A}{{\bf A}}

\newcommand{\tB}{{\widetilde{\bf B}}}
\newcommand{\tA}{{\widetilde{\bf A}}}
\newcommand{\tx}{{\widetilde{\bf x}}}
\newcommand{\I}{{\bf I}}
\newcommand{\x}{{\bf x}}

\newcommand{\e}{{\bf e}}

\title{Estimation of linear, non-gaussian causal models\\ in the presence
of confounding latent variables}
\author{Patrik O.\ Hoyer$^1$, Shohei~Shimizu$^{2,1}$ \and Antti J.\ Kerminen$^1$\\ 
\parbox[t]{75mm}{
\center
$^{1)}$ HIIT Basic Research Unit\\  
Department of Computer Science\\ 
University of Helsinki\\
Finland} 
\parbox[t]{75mm}{
\center
$^{2)}$ Learning and Inference Group\\
The Institute of Statistical Mathematics\\
Japan
}
}

\summary{%
The estimation of linear causal models (also known as structural equation models) from data is a well-known problem which has received much attention in the past. Most previous work has, however, made an explicit or implicit assumption of gaussianity, limiting the identifiability of the models. We have recently shown \cite{Shimizu05uai,Hoyer06ica} that for non-gaussian distributions the full causal model can be estimated in the no hidden variables case. In this contribution, we discuss the estimation of the model when confounding latent variables are present. Although in this case uniqueness is no longer guaranteed, there is at most a finite set of models which can fit the data. We develop an algorithm for estimating this set, and describe numerical simulations which confirm the theoretical arguments and demonstrate the practical viability of the approach. Full Matlab code is provided for all simulations.
}

\begin{document}
\maketitle

\renewcommand{\thepage}{}
\thispagestyle{myheadings}
\markright{{\upshape [Submitted draft. See {\ttfamily www.cs.helsinki.fi/patrik.hoyer/} for latest version and citation info.]}}

%This is a submitted draft. Please see\\ 
%http://www.cs.helsinki.fi/patrik.hoyer/ \\
%for the latest version and citation information. \\

\section{Introduction}

The ultimate goal of much of the empirical sciences is the discovery of \emph{causal relations}, as opposed to just correlations, between variables. That is, one is not only interested in making predictions based on observations; one also wants to predict what will happen if one intervenes and changes something in the system. 

In many cases, causal effects can be estimated using controlled randomized experiments. Unfortunately, however, in many fields of science and in many studies it is not always possible to perform controlled experiments. Often it can be very costly, unethical, or even technically impossible to directly control the variables whose causal effects we wish to learn. In such cases one must rely on observational studies combined with prior information and reasonable assumptions to learn causal relationships. This has been called the \emph{causal discovery} problem \cite{Pearl00book,Spirtes00book}.

Linear causal models, also known as structural equation models, can be thought of as the simplest possible causal models for continuous-valued data, and they have been the target of much research over the past decades, see e.g.\ \cite{Bollen89book,Silva06} and references therein. The bulk of this research has, however, made an implicit assumption of gaussian data (i.e.\ normally distributed data). Although the methods that have been developed work for any distributions, they have been limited in their estimation abilities by what can be accomplished for gaussian data. Fortunately, in the real world, many variables are inherently non-gaussian, and in such a case one can obtain much stronger results than for the gaussian case. In earlier work \cite{Shimizu05uai,Hoyer06ica} we showed that if the variables involved are non-normally distributed, yet linearly related to each other, the complete causal graph is identifiable from non-experimental data if no confounding hidden variables are present. In the present paper we show what can be done if this last assumption is violated. Although in this case uniqueness is no longer guaranteed, there is at most a finite set of models which can fit the data. 

This paper is organized as follows. First, in sections~\ref{sec:lin} and \ref{sec:can}, we define the linear causal models that are the focus of this study. Section~\ref{sec:algo} describes how to estimate the causal model from data. Section~\ref{sec:simulations} provides some simulations confirming the practical viability of the approach. Finally, sections~\ref{sec:future} and \ref{sec:conc} discuss future work and state some conclusions.

\begin{figure*}
\centerline{
\resizebox{0.9 \textwidth}{!}{
\includegraphics{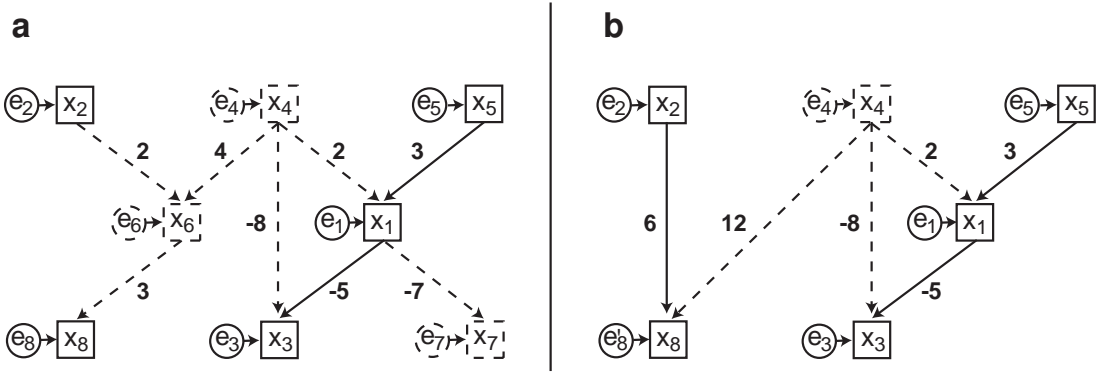}}}
\caption{{\bf (a)} An example of a latent variable LiNGAM model. The diagram corresponds to the following data generation process: $x_2:=e_2$, $x_4:=e_4$, $x_5:=e_5$, $x_6:=2x_2+4x_4 + e_6$, $x_1 := 2x_4 + 3x_5 + e_1$, $x_8 := 3x_6+e_8$, $x_3:=-8x_4 -5x_1 + e_3$, and $x_7 := -7x_1+e_7$. (Here, all the $c_i$ are zero, but this is not the case in general.) The `disturbance' variables $e_i$ are drawn mutually independently from \emph{non-gaussian} distributions $p_i(e_i)$. Hidden variables are shown dashed; the observed data vector $\x$ consists of only the values of $x_1$, $x_2$, $x_3$, $x_5$, and $x_8$. {\bf (b)} The canonical model corresponding to the network in (a). This is the observationally and causally equivalent network where all causally irrelevant variables have been simplified away. See main text for details.
\label{fig:exlvnetwork}}
\end{figure*}

\section{Linear causal models} 
\label{sec:lin} 
  
Assume that we observe data generated by a linear, non-gaussian, acyclic model (i.e.\ a LiNGAM-process \cite{Shimizu05uai,Hoyer06ica}) but that we only observe a subset of the involved variables. That is, the process has the following properties:
\begin{enumerate}
\item The full set of variables (including unobserved variables) $x_i, \; i=\{1 \ldots m\}$ can be arranged in a \emph{causal order}, such that no later variable causes any earlier variable. We denote such a causal order by $k(i)$. That is, the generating process is \emph{recursive} \cite{Bollen89book}, meaning it can be represented by a \emph{directed acyclic graph} (DAG) \cite{Pearl00book,Spirtes00book}. 
\item The value assigned to each variable $x_i$ is a \emph{linear function} of the values already assigned to the earlier variables, plus a `disturbance' (error) term $e_i$, and plus an optional constant term $c_i$,  that is
\begin{equation}
x_i = \sum_{k(j)<k(i)} \tilde{b}_{ij} x_j + e_i + c_i.
\end{equation}
\item The disturbances $e_i$ are all zero-mean continuous random variables with \emph{non-gaussian} distributions of non-zero variances, and the $e_i$ are independent of each other, i.e.\ $p(e_1, \ldots, e_m) = \prod_i p_i(e_i)$.
\item The \emph{observed variables} is a subset of the $x_i$. We denote the set containing the indices of the observed variables by $J \subseteq \{1, \ldots, m\}$. In other words, our data set contains only the $x_j, \; j \in J$.
\end{enumerate}
Figure~\ref{fig:exlvnetwork}a shows an example of such a \emph{latent variable LiNGAM model}. 

An additional assumption not needed in our earlier work \cite{Shimizu05uai,Hoyer06ica} but useful for the purposes of this paper is the requirement that the generating network is \emph{stable} \cite{Pearl00book} such that there is no exact canceling of effects. That is, if there is a directed path from $x_i$ to $x_j$ then $x_i$ has a causal effect on $x_j$. This condition has also been termed \emph{faithfulness} \cite{Spirtes00book}, and in our context implies that when multiple causal paths exist from one variable to another their combined effect does not equal exactly zero.

Finally, we assume that we are able to observe a large number of data vectors $\x$ (which contain the observed variables $x_j, \; j\in J$), and that each is generated according to the above described process, with the same set of observed variables $J$, same causal order $k(i)$, same coefficients $\tilde{b}_{ij}$, same constants $c_i$, and the disturbances $e_i$ sampled independently from the same distributions. 

The key difference to our previous work \cite{Shimizu05uai,Hoyer06ica} is that we here allow \emph{unobserved confounding variables}: hidden variables which affect multiple observed variables and hence are potentially problematic for any causal analysis. The key difference to other existing research involving linear models with latent variables, such as \cite{Bollen89book,Silva06}, is our assumption of \emph{non-gaussian} variables, allowing us to utilize methods based on higher-order statistics.

\section{Canonical models}
\label{sec:can} 

Consider the example model shown in Figure~\ref{fig:exlvnetwork}a. It should be immediately clear that it is impossible to estimate the full generating model from a sample of data vectors $\x = (x_1, x_2, x_3, x_5, x_8)^T$. This can be seen most clearly by the fact that the data generating model contains a hidden variable ($x_7$) with no observed descendants; detecting the presence of such a variable from our data is obviously not feasible since it has absolutely no effect on the observed data. Fortunately, the impossibility of detecting $x_7$ and of estimating the strengths of any connections to it are not cause for concern. The reason for this is that the variable is not in any way relevant with respect to our goal: finding the causal relationships between the observed variables. 

Another causally irrelevant hidden variable is $x_6$, which simply mediates the influence of $x_2$ and $x_4$ onto $x_8$. We have
\begin{eqnarray}
x_6 & := & 2x_2 + 4x_4 + e_6 \nonumber \\
x_8 & := & 3x_6 + e_8 \nonumber
\end{eqnarray}
leading to
\begin{eqnarray}
x_8 & := & 3(2x_2 + 4x_4 + e_6) + e_8 \nonumber \\
& = & 6x_2 +12x_4 +3e_6+e_8 \nonumber \\
& = & 6x_2 +12x_4 +e'_8 \nonumber
\end{eqnarray}
where we have simplified $e'_8 = 3e_6+e_8$. This shows that we can remove the hidden variable $x_6$ from the model to obtain a simpler model in which the observed data is identical to the original one and, in addition, the causal implications are the same. The resulting model is shown in Figure~\ref{fig:exlvnetwork}b. Note that hidden variable $x_4$ cannot be simplified away without changing the observed data.

We formalize this idea of causally relevant versus irrelevant hidden variables using the following concepts:

Two latent variable LiNGAM models are \emph{observationally equivalent} if and only if the distribution $p(\x)$ of the observed data vector is identical for the two models. This implies that the models cannot be distinguished based on observational data alone.

Two latent variable LiNGAM models are \emph{observationally and causally equivalent} if and only if they are observationally equivalent and all causal effects of observed variables onto other observed variables are identical for the two models. In the notation of \newcite{Pearl00book} these causal effects are given by $p(\x_{J_1}\; |\; \doo( \x_{J_2} )),\; J_1,J_2 \subseteq J,$ with $J_1 \cap J_2 = \emptyset$. When these are identical for all choices of $J_1$ and $J_2$ the two models in question cannot be distinguished based on any observations nor any controlled experiments.

Finally, we define a \emph{canonical model} to be any latent variable LiNGAM model where each latent variable is a root node (i.e.\ has no parents) and has at least two children (direct descendants). Furthermore, although different latent variables may have the same sets of children, no two latent variables exhibit exactly proportional sets of connection strenghts to the observed variables. Finally, each latent variable is restricted to have zero mean and unit variance.

To derive a canonical model which is observationally and causally equivalent to any given latent variable LiNGAM model, we can use the following algorithm:

\hspace{-1.3\parindent}
\rule{\columnwidth}{0.5mm}
{ \sffamily
{\bf Algorithm A:}
Given a latent variable LiNGAM model, returns an observationally and causally equivalent canonical model
\begin{enumerate}
\item \label{step:removedeadends} First remove any latent variables without children. Iterate this rule until there are no more such nodes.
\item \label{step:mediating} For any connection of the form $X \rightarrow Y$, where $Y$ is a latent variable: (a) For all children $Z$ of $Y$, add a direct connection $X \rightarrow Z$, the strength of the connection being the product of the strenghts of $X \rightarrow Y$ and $Y \rightarrow Z$. If a direct connection already existed, add the specified amount to the original strength. (b) Remove the connection $X \rightarrow Y$. Iterate this rule until there are no more applicable connections.
\item \label{step:removesingles} Remove any latent variable with only a single child, incorporating the latent variable's disturbance variable and constant into those of the child. Iterate this until there are no more such latent variables.
\item For any pair of latent variables with \emph{exactly proportional} sets of connection strenghts to the observed variables, combine these into a single latent variable. Iterate this until there are no more such pairs of latent variables.
\item Finally, standardize the latent variables to have zero mean and unit variance by adjusting the connection strenghts to, and the constants of, their children.
\end{enumerate}
} \vspace{-3.5mm}
\hspace{-1.3\parindent}
\rule{\columnwidth}{0.5mm}

\section{Model estimation}
\label{sec:algo}

\begin{figure*}
\centerline{
\resizebox{0.9 \textwidth}{!}{
\includegraphics{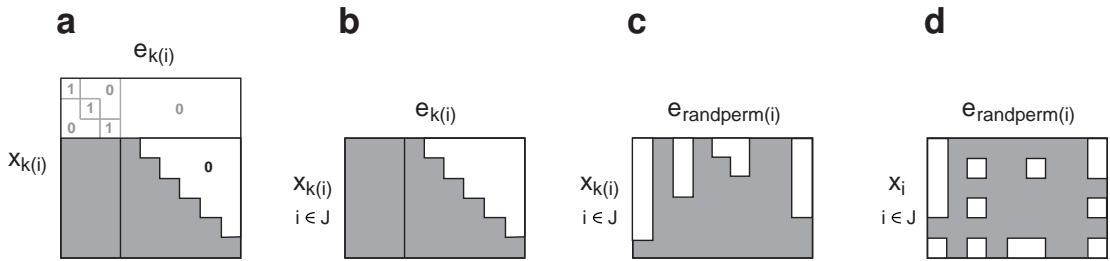}}}
\caption{{\bf (a)} Basic structure of the full ICA matrix $\tA$ for a canonical model. (In this example, there are 3 hidden and 6 observed variables.) The top rows correspond to the hidden variables. Note that the assumption of a canonical model implies that this part of the matrix consists of an identity submatrix and zeros. The bottom rows correspond to the observed variables. Here, the shaded area represents values which, depending on the network, \emph{may be zero or non-zero}. The white area represents entries which are zero by the DAG-assumption. {\bf (b)} Since, by definition, we do not observe the latent variables, the information that we have is limited to the bottom part of the matrix shown in (a). {\bf (c)} Due to the permutation indeterminancy in ICA, the observed basis matrix has its columns in random order. {\bf (d)} Since we do not (up front) know a causal order for the observed variables, the observed mixing matrix $\A$ has the rows in the order of data input, not in a causal order.
\label{fig:Astructure}}
\end{figure*}

In this section we show how, from data generated by any stable latent variable LiNGAM model, to estimate the set of canonical models which are observationally equivalent to the generating model. 

\subsection{Model estimation by ICA}

In earlier work \cite{Shimizu05uai,Hoyer06ica} we showed that data generated from a LiNGAM process follows an independent component analysis (ICA) distribution \cite{Comon94,Hyva01book}. Here, we briefly review this concept, specifically focusing on the effect of latent variables. 

We begin by considering the full data vector $\tx = \{x_1, \dots, x_m\}$, which includes the latent variables. If we as preprocessing subtract out the means of the variables, then the full data satisfies $\tx = \tB\tx + \e,$ where, because of the DAG assumption, $\tB$ is a matrix that could be permuted  to strict lower triangularity if one knew a causal ordering $k(i)$ of the variables. Solving for $\tx$ one obtains $\tx = \tA\e,$ where $\tA = (\I-\tB)^{-1}$ contains the influence of the disturbance variables onto the observed variables. Again, $\tA$ could be permuted to lower triangularity (although not \emph{strict} lower triangularity) with an appropriate permutation $k(i)$. Taken together, the linear relationship between $\e$ and $\tx$ and the independence and non-gaussianity of the components of $\e$ define the standard linear \emph{independent component analysis} model.

So far, this is just restating what we pointed out in our previous work. Now consider the effect of hiding some of the variables. This yields $\x = \A\e,$ where $\A$ contains just the rows of $\tA$ corresponding to the observed variables. When the number of observed variables is less than the number of disturbance variables, $\A$ is non-square with more columns than rows. This is known as an \emph{overcomplete} basis in the ICA literature.

Let us take a closer look at the structure inherent in the `mixing matrix' $\A$. First, note that since for every latent variable LiNGAM model there is an observationally and causally equivalent canonical model, we can without loss of generality restrict our analysis to canonical models. Next, arrange the full set of variables such that all latent variables come first (in any internal order) followed by all observed variables (in a causal order), and look at the structure of the full matrix $\tA$ shown in Figure~\ref{fig:Astructure}a. Although we only observe part of the full matrix, with randomly permuted columns and with arbitrarily permuted rows as in Figure~\ref{fig:Astructure}d, the crucial point is that the observed ICA basis matrix $\A$ contains all of the information contained in the full matrix $\tA$ in the sense that all the free parameters in $\tA$ are also contained in $\A$. Thus there is hope that the causal model could be reconstructed from the observed basis matrix $\A$.

At this point, we note that we of course do not directly observe $\A$ but must infer it from the sample vectors. \newcite{Eriksson04} have recently shown that the overcomplete ICA model is identifiable given enough data, and several algorithms are available for this task, see e.g.\  \cite{Moulines97,Attias99,Hyva01book}.
Thus, the remainder of this subsection considers the inference of the causal model were the exact ICA mixing matrix $\A$ known. The next subsection deals with the practical aspect of dealing with inevitable estimation errors.

As in the standard square ICA case, identification in the overcomplete case is only up to permutation and scaling of the columns of $\A$. The scaling indeterminancy is not serious; it simply amounts to a problem of not being able to attribute the magnitude of the influence of a disturbance variable $e_i$ to its variance, the strength of its connection to its corresponding variable $x_i$, and in the case of hidden variables the average strength of the connections from that hidden variable to the observed variables. This is of no consequence since we are anyway never able to directly monitor the hidden variables nor the disturbance variables, making the scaling simply a matter of definition.

In contrast, the permutation indeterminancy \emph{is} a serious problem, and in general leads to non-uniqueness in inferring the model: We cannot know which columns of $\A$ correspond to the hidden variables. Note, however, that this is the only information missing, as illustrated in Figure~\ref{fig:Astructure}. Thus, an upper bound for the number of observationally equivalent canonical models is the number of classifications into observed vs hidden. This is simply $(N_o+N_h)!/(N_o! N_h!),$ where $N_o$ and $N_h$ denote the numbers of observed and hidden variables. For each classification we need to check whether this leads to a valid latent-variable LiNGAM model:

\hspace{-1.3\parindent}
\rule{\columnwidth}{0.5mm}
{ \sffamily
{\bf Algorithm B:}
Given an overcomplete basis $\A$ (containing exact zeros) and the means of the observed variables, calculates all observationally equivalent canonical latent variable LiNGAM models compatible with the basis
\begin{enumerate}
\item \label{step:determineNh} $N_h$ is determined as the number of columns of $\A$ minus the number of rows. 
\item \label{step:iterateallcomb} For each possible classification of the columns of $\A$ as belonging to disturbance variables of observed vs hidden variables:
\begin{enumerate}
\item Reorder the columns such that the ones selected as `hidden variables' come first
\item Augment the basis by adding the unobserved top part of the matrix (as in Figure~\ref{fig:Astructure}a), obtaining an estimate of $\tA$.
\item Test whether it is possible to permute $\tA$ (using independent row and column permutations) to lower-triangular form. If not, go to the next classification.
\item Divide each column of $\tA$ by its diagonal element, and calculate the connection strengths $\tB = \I - \tA^{-1}$
\item Check that the found network is compatible with the \emph{stability} assumption. If not, go to the next classification.
\item Add the found network $\tB$ to the list of observationally equivalent models which could have generated the data.
\end{enumerate}
\end{enumerate}
} \vspace{-4mm}
\hspace{-1.3\parindent}
\rule{\columnwidth}{0.5mm}

\subsection{Practical considerations}
\label{sec:practicalities}

Up to here we have assumed that the exact ICA basis matrix is known. In a real implementation, of course, it can only be \emph{estimated} from the data, inevitably leading to (small) estimation errors, causing all elements of $\A$ to be non-zero and making Algorithm B not directly applicable. Furthermore, since the structure of the network is related to the \emph{inverse} of $\A$, small estimation errors will give rise to small direct effects where none exist in the generating model.

Solving these problems requires us to have an idea of the accuracy of our estimate of $\A$. Here, we advocate using resampling methods \cite{Efron93book}, obtaining a \emph{set} of estimates $\A_i$ representing our uncertainty regarding the elements of the mixing matrix. This set can then be used as follows: First, one can infer which elements of $\A$ are exactly zero by standard statistical testing, taking as the null hypothesis the assumption that a given element is zero and rejecting it when the mean/variance ratio for that element is large enough (in an absolute value sense). This procedure will give the correct answer with probability approaching 1 as the amount of data grows. 

Having identified the zeros of the basis matrix, we apply Algorithm B to each estimate $\A_i$ to yield a set of estimates for the generating causal model.\footnote{Note that this is, in general, a set of sets: One index corresponds to the different possible permutations of the columns of $\A$, the other to the different estimates by resampling.}  This set can, again, be utilized in statistical tests to prune out the small direct effects which result from estimation errors. 

A full description of the above ideas is out of the scope of this short paper. Nevertheless, they have been successfully implemented in our simulations (see Section~\ref{sec:simulations}) and the reader is invited to study the code package for all the details. 

\section{Simulations}
\label{sec:simulations}
 
\begin{figure*}
\centerline{
\resizebox{0.9 \textwidth}{!}{
\includegraphics{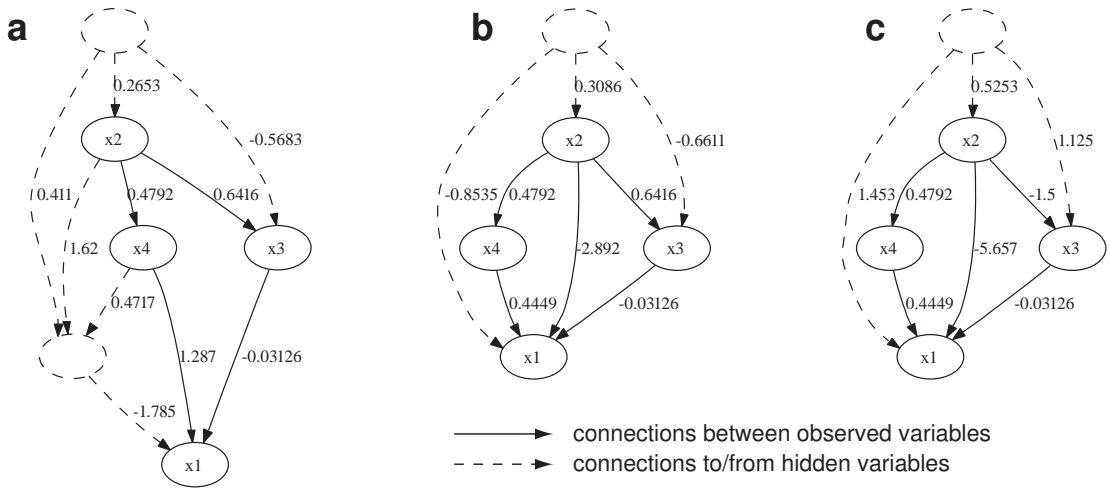}}}
\caption{{\bf (a)} A randomly generated latent variable LiNGAM model. {\bf (b)} The canonical model which is causally and observationally equivalent to the network in (a). {\bf (c)} An observationally equivalent (but causally \emph{not} equivalent) model. From the ICA basis which model (a) generates, it is impossible to distinguish between models (b) and (c). 
See main text for details.
\label{fig:tests}}
\end{figure*}

We have performed extensive simulations in order to (i) verify the algorithms described, (ii) test the viability of the resampling approach for dealing with estimation errors, and (iii) provide a simple demonstration of learning a small hidden variable LiNGAM model from data. Full well-documented Matlab code\footnote{at {\ttfamily http://www.cs.helsinki.fi/patrik.hoyer/\\ \parbox{44.5mm}{\ } code/lvlingam.tar.gz}} for all of these experiments is available, to allow the reader to effortlessly replicate and verify our results, which unfortunately cannot be described in very great detail here due to lack of space.

First, we tested the basic idea by generating random latent variable LiNGAM models and, for each model, calculating its corresponding ICA basis matrix $\A$ and from it inferring the set of observationally equivalent canonical models using Algorithm B. This process is illustrated in Figure~\ref{fig:tests}. In every case, exactly one model in the inferred set was causally equivalent to the original model.

Second, we tested the practical viability of the approach by assuming that we only have a set of inexact estimates of the basis matrix: We generated random LiNGAM models and, for each model, calculated its corresponding ICA basis matrix, added noise to yield 20 different `estimates' of it, and used the approach of Section~\ref{sec:practicalities} to infer the set of possible generating models. In each case, one of the inferred models was close (in a causal sense) to the original model. When the noise becomes large enough to cause misclassifications of zeros, the method can fail and the resulting networks are not necessarily close to the generating one. This is analogous to the sensitivity of constraint-based causal discovery algorithms \cite{Pearl00book,Spirtes00book} to misclassifications of conditional independencies in the observed data. 

\begin{figure}
\centerline{
\resizebox{0.9 \columnwidth}{!}{
\includegraphics{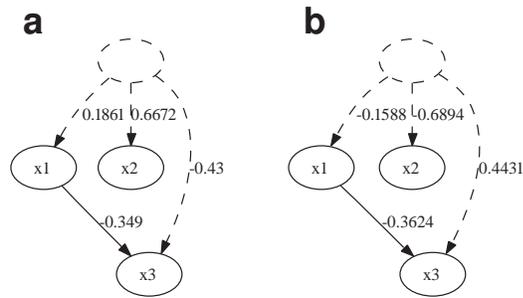}}}
\caption{{\bf (a)} A randomly generated latent variable LiNGAM model used to generate data. {\bf (b)} The model estimated from the data. Note that we cannot determine the sign of the influence of the hidden variable since its distribution was symmetric. All other parameters are correctly identified, including the absence of connections between $x_1$ and $x_2$, and $x_2$ and $x_3$. 
\label{fig:demo}}
\end{figure}

Finally, we performed a small-scale demonstration of learning the model from actual data. We used the mixture-of-gaussians framework \cite{Moulines97,Attias99} to estimate the ICA bases. To make the problem as simple as possible, the simulation was done on very small networks (3 observed and one hidden variable) and we additionally assumed that we knew the distributions of all the disturbance variables. A sample of 1000 data vectors was used. Figure~\ref{fig:demo} shows a typical result. The algorithm here finds the correct structure of the network (in this case the network is uniquely identifiable). This would not be possible with conventional methods as there are no conditional independencies among the observed variables. 

\section{Future work}
\label{sec:future}

Much work remains before the framework described here can be applied to practical data analysis problems. The most formidable problem is that of reliably estimating an overcomplete ICA basis matrix when the number of hidden variables and the distributions of the disturbance variables are unknown. Current algorithms \cite{Moulines97,Attias99} may work for very small dimensions but to our knowledge no methods are yet available for estimating overcomplete ICA bases reliably and accurately in high dimensions.

Fortunately, we may be able to solve relatively large problems even with current methods. The key is that hidden variables only cause what might be called \emph{local overcompleteness}. In a large LiNGAM model where each latent variable directly influences only a small number of observed variables, the data follows an independent subspace (ISA) model \cite{Hyva00NC_complex}. In such a model, the data distribution can be represented as a combination of low-dimensional independent subspaces. When the data follows the ISA model, the individual subspaces might still be found using efficient ICA algorithms \cite{Hyva01book}, and algorithms for overcomplete ICA would only need to be run \emph{inside each subspace}. 

\section{Conclusions}
\label{sec:conc} 

We have recently shown how to estimate linear causal models when the distributions involved are non-gaussian and no confounding hidden variables exist \cite{Shimizu05uai,Hoyer06ica}. In this contribution, we have described the effects of confounding latent variables, and shown how to estimate the set of models consistent with the observed data. 
Simulations, for which full Matlab code is available, confirm the viability of the approach. Further work is needed to develop the software into a practical, easy-to-use data-analysis tool.

\subsection*{Acknowledgments}

We thank Aapo Hyv\"arinen for comments on\newpage \noindent the manuscript. P.O.H.\ was supported by the Academy of Finland project \#204826.

\bibliographystyle{acl}
\bibliography{/users/phoyer/work/bib/bibfiles/collection,/users/phoyer/work/bib/bibfiles/personal}

\end{document}